\documentclass[]{cas-dc} 

\usepackage{bm}
\usepackage{caption,subcaption}  

\usepackage[square,numbers]{natbib}

\def\tsc#1{\csdef{#1}{\textsc{\lowercase{#1}}\xspace}}
\tsc{WGM}
\tsc{QE}

\begin{document}
\let\WriteBookmarks\relax
\def\floatpagepagefraction{1}
\def\textpagefraction{.001}

\shorttitle{Iterative GP-MLMPC for (Semi-)Batch Processes}    

\shortauthors{T.X.Tan et al.} 

\title [mode = title]{Iterative Model-Learning Scheme via Gaussian Processes for Nonlinear Model Predictive Control of (Semi-)Batch Processes}  

\author[1]{Tai Xuan Tan}

\credit{Conceptualization, Methodology, Software, Formal Analysis, Investigation, Writing - original draft, Visualization}

\affiliation[1]{organization={Process Systems Engineering (AVT.SVT), RWTH Aachen University},
            addressline={Forckenbeckstraße 51}, 
            city={52074 Aachen},
            country={Germany}}

\author[2,1,3]{Alexander Mitsos}

\credit{Conceptualization, Writing - review \& editing, Supervision, Funding Acquisition}

\affiliation[2]{organization={JARA-CSD},
            addressline={52056 Aachen}, 
            country={Germany}}

\affiliation[3]{organization={Institute of Climate and Energy Systems: Energy Systems Engineering (ICE-1), Forschungszentrum Jülich GmbH},
            addressline={52425 Jülich}, 
            country={Germany}}

\author[4]{Eike Cramer}[orcid = 0000-0002-6882-5469]
\cormark[1]

\credit{Conceptualization, Writing - original draft, Supervision, Funding Acquisition}

\affiliation[4]{organization={Department of Chemical Engineering, Sargent Centre for Process Systems Engineering,
University College London},
            addressline={Torrington Place, 
London WC1E 7JE},
country = {United Kingdom}}

\cortext[cor1]{Corresponding author: e.cramer@ucl.ac.uk}

\begin{abstract}
Batch processes are inherently transient and typically nonlinear, motivating nonlinear model predictive control (NMPC).
However, adopting NMPC is hindered by the cost and unavailability of dynamic models.
Thus, we propose to use Gaussian Processes (GP) in a model-learning NMPC scheme (GP-MLMPC) for batch processes. 
We initialize the GP-MLMPC using data from a single initial trajectory, e.g., from a PI controller.
We iteratively apply the NMPC embedded with GPs to run batches and update the GP with new observations from each iteration, thereby achieving batch-wise improvements.
Using uncertainty quantification from the GPs, we formulate chance constraints to enforce safe operation to the required confidence levels.
We demonstrate our approach in \textit{silico} on a semi-batch polymerization reactor for tracking and economic objectives over durations of two hours, and the reactor temperature is constrained in a range of $\pm2^\circ C$ around its setpoint.
After only four batch iterations, tracking error from the GP-MLMPC scheme converged to a reduction of $83\%$, compared to the initial trajectory.
Furthermore, under an economic objective, the GP-MLMPC resulted in a 17-fold increase in final product mass by iteration 8, compared to the initial trajectory.
In both cases, the resulting GP-MLMPC performance is on par with the full-model NMPC, which shows that the optimal controller can be learned by the approach. 
By collecting samples around the optimal trajectory, the GP-MLMPC remains sample-efficient across iterations and achieves quick convergence.
Thus, the proposed GP-MLMPC scheme presents a promising data-efficient approach for the control of nonlinear batch processes without mechanistic knowledge. 
\end{abstract}

\begin{keywords}
 Learning-based nonlinear control\sep 
 Batch Process Control\sep
 Nonlinear Model Predictive Control\sep
 Machine Learning\sep 
 Gaussian Processes\sep
 Chance Constraints\sep
\end{keywords}

\maketitle

\section{Introduction}\label{sec: Introduction} 
Batch processes with extensions to semi-batch remain one of the primary manufacturing processes for products in small to medium quantities, as well as for diverse consumer specifications~\cite{Yoo2021}, e.g., specialty polymers and pharmaceuticals. 
Batch processes are transient in operation, and therefore are exposed to nonlinearities in the system dynamics, as there are no steady operating points for control design to linearize around~\cite{Lee2006}.
Therefore, conventional linear optimal controllers cannot be easily utilized for satisfactory control of batch processes.
For specialty products, these processes also often have to adhere to tight process constraints corresponding to narrow ranges of quality requirements~\cite{Yoo2021}.
To address the aforementioned challenges, batch processes require control methods that differ from those used in continuous, steady-state processes.

(Nonlinear) Model predictive control ((N)MPC) is a well-established advanced feedback control method~\cite{Rawlings2017} and has been extensively studied for many constrained processes, including batch process systems~\cite{Lauri2014, Lucia2014}.
Detailed explanation of (N)MPC can be found here~\cite{Rawlings2017}.
(N)MPC relies on an accurate dynamic model of the system to perform well.
However, developing accurate mechanistic models and experimentally validating these can be prohibitively expensive, especially for batch units that produce a wide range of products, each directly affecting the dynamic behavior of the process.
The cost of obtaining a model for MPC is not often justifiable for batch processes, which typically produce small quantities of products.
Thus, the lack of available models for control is one of the main limiting factors for the wide adoption of NMPC in the batch process industry~\cite{Fiedler2023}.

Data-driven models have the potential to overcome this challenge by acting as surrogate models of the system dynamics. For batch processes, Marquez-Ruiz et al.~\cite{Marquez-Ruiz2019} proposed an alternative approach named model learning predictive control.
The method features an iterative scheme where the linear parameter varying models (LPV) used for control are updated between batches using new measurement data from performed batches.
The approach demonstrated robust performance in the time domain and showed significant improvement along the batch iteration domain.
However, LPV models depend on a measurable scheduling variable, which might not be clearly defined for all systems.
It can also occur that the nonlinear dynamics from complex reactions cannot be sufficiently described by LPV models.
Therefore, this work aims to extend the learning-based MPC scheme to NMPC controllers by using flexible nonlinear surrogate models.

Gaussian Processes (GPs) regression is a class of probabilistic universal approximators~\cite{Rasmussen2006}.
Commonly used for surrogate modeling, GPs have been applied to many different applications, including NMPC~\cite{Murray-Smith2003, Maciejowski2013, Klenske2016, Bradford2020, Hewing2020a, Maiworm2021}. 
The GPs' probabilistic regression quantifies the model uncertainty associated with the predictions given available data, which can be utilized in the NMPC framework to identify regions of confidence.
The earliest example of GP based NMPC (GP-NMPC) was introduced by Murray-Smith~et~al.~\cite{Murray-Smith2003} for reference tracking problems without constraints. 
Bradford~et~al.~\cite{Bradford2020} utilized the mean and variance functions from GPs to implement stochastic NMPC, where back-off constants for chance constraints were precomputed via Monte Carlo samples of the offline GPs. 
Hewing~et~al.~\cite{Hewing2020a} used GPs to account for residuals given a nominal model in constrained NMPC for autonomous vehicles. 
For GP updates, Maiworm~et~al.~\cite{Maiworm2021} implemented online GP updates with system measurements for model predictive control, with a simulated continuous reactor.
The aforementioned works illustrate the potential of GP-NMPC in learning-based control.
However, there is a literature gap where most works only consider offline learning of the GP model before using it for NMPC.
And while Maiworm~et~al.~\cite{Maiworm2021} showed GP updates, the initial dataset in the work was designed to cover the state-action space evenly, with excitation signals.
In realistic scenarios, data is often scarce and comes from control trajectories, which often lie clustered in the state-action space domain.
Hence, we study iterative updates of GP-NMPC where only small initial data are available, instead of relying on a designed initial dataset covering the operating region.

We propose a model learning scheme for GP-NMPC (GP-MLMPC) in batch processes, where the GP model is updated iteratively after every completed batch, with noisy measurements of the system, as inspired by the ML-MPC scheme~\cite{Marquez-Ruiz2019}.
By the proposed iterative scheme, the GP-NMPC collects measurements mostly towards and around the optimal trajectory.
As the surrogate model only requires accurate predictions around the optimal trajectory for control, the scheme provides a sample-efficient method of learning for the GP model.
As opposed to existing works on GP-NMPC, we study the iterative performance of the controller after each update.
Furthermore, we implement chance constraints with uncertainty quantification from the GP surrogate model.
When system measurements are limited, the resulting surrogate model can have large uncertainties in certain operating regions.
The chance constraints restrict the likelihood of constraint violation by selecting safer actions, thus probabilistically ensuring constraint satisfaction.
In summary, our contribution  is three-fold:
First, we propose the GP-MLMPC scheme for batch processes as a framework for NMPC without an existing model.
Second, we show empirically that the GP-MLMPC control performance converges quickly towards the full-model NMPC for both tracking and economic objectives, indicating high data-efficiency.
Lastly, we demonstrate the effect of chance constraints for safe learning and operation of uncertain systems.

The article is comprised of the following sections. Section~\ref{sec: GP_state_space} describes GP regression for state space modeling, including the sparse GP variant and multi-step ahead predictions.
In Section~\ref{sec: GP-NMPC}, we detail the implementation of NMPC with fitted GP as a surrogate model. 
In Section~\ref{sec: learning scheme}, we elaborate on the GP-MLMPC scheme to iteratively improve the control performance by using GP updates. 
In Section~\ref{sec: case study and results}, we describe the semi-batch polymerization reactor case study and discuss the results from applying the proposed learning method. 
Finally, Section~\ref{sec: conclusion} concludes this work and points to future directions.

\section{Gaussian Process-based State Space Modeling}\label{sec: GP_state_space}
We briefly introduce the Gaussian Process (GP) and its sparse variant. 
We then describe the use of GPs for state space modeling and outline multi-step ahead predictions.

\subsection{Gaussian Processes}\label{ssec: GP}
We use Gaussian Processes for probabilistic nonlinear regression, i.e., GPs are applied to learn nonlinear latent functions $f:\mathbb{R}^{d_{\bm{z}}}\rightarrow\mathbb{R}$.
For a more complete description of GPs, refer to~\cite{Rasmussen2006}.
Each noisy observation $y \in \mathbb{R}$ is given by:
\begin{equation*}
    y = f(\bm{z}) + \epsilon
\end{equation*}
Here, $\bm{z} \in \mathbb{R}^{d_{\bm{z}}}$ is the function vector input, while the Gaussian noise $\epsilon\sim N(0,\sigma_\epsilon^2)$. 
We denote the dataset as $\mathcal{D} = \{\bm{Y},\bm{Z}\}$, where $\bm{Y} = [y_1,y_2,...,y_{n_D}]^T$, and $\bm{Z} = [\bm{z}_1,\bm{z}_2,...,\bm{z}_{n_D}]^T$, i.e., the vector of $n_D$ noisy measurement data and the corresponding matrix of input data, respectively. 

By conditioning the GP on dataset $\mathcal{D}$, we analytically obtain the posterior distribution $p(f(\bm{z})|\mathcal{D})$.
These are the main GP equations used to infer Gaussian predictions:
\begin{equation}\label{eq: Gaussian process}
    \begin{aligned}
        p(f(\bm{z})|\mathcal{D}) &\sim \mathcal{N}(\mu_f (\bm{z}), \sigma^2_f(\bm{z}))\\
        \mu_f(\bm{z}) &= \bm{k}^T(\bm{z}) (\bm{K} + \sigma_n^2\bm{I})^{-1} \bm{Y}\\
        \sigma^2_f(\bm{z}) &=  k(\bm{z},\bm{z})- \bm{k}(\bm{z})^T(\bm{K}+\sigma_n^2\bm{I})^{-1}\bm{k}(\bm{z}) 
   \end{aligned}
\end{equation}
The vector $\bm{k} =[k(\bm{z},\bm{z}_1),..., k(\bm{z},\bm{z}_{n_D})]^T\in \mathbb{R}^{n_D}$ contains the covariances between the prediction point and the training points, 
while the symmetric covariance matrix $\bm{K} \in \mathbb{R}^{n_D\times n_D} $ contains the covariances between the training points, i.e., the element $\bm{K}_{ij} = k(\bm{z}_i,\bm{z}_j)$ and $\bm{z}_i,\bm{z}_j \in \bm{Z}$. 
The hyperparameter $\sigma_n^2$ represents the estimated variance of measurement noise. 

A GP is fully specified by a prior mean function, $m(\bm{z})$, and a covariance function, $k(\bm{z}_i,\bm{z}_j)$.
We implement the prior mean $m(\bm{z}) = 0$ for a standardized dataset without loss of generality.
For the covariance function, we implement the popular squared exponential (SE) function, due to its flexibility and universal approximation property~\cite{Micchelli2006}.
\begin{equation*}\label{eq: SE function}
    k(\bm{z}_i,\bm{z}_j) = \sigma_f^2  \exp\left(-\frac{1}{2}(\bm{z}_i-\bm{z}_j)^T\bm{\Lambda}^{-1}(\bm{z}_i-\bm{z}_j)\right)
\end{equation*}
The diagonal matrix $\bm{\Lambda} = \text{diag}(\ell_1^2,\ell_2^2,...,\ell^2_{d_z})$ contains the squared length scale hyperparameter for each input dimension.
The hyperparameter $\sigma_f^2$ is defined as the so-called signal variance. The hyperparameter set $\Theta =\{\sigma_n^2,\sigma_f^2,\bm{\Lambda}\}$ is optimized to maximize the log marginal likelihood, i.e., probability of GP describing the given dataset.
Please refer to Appendix A for the expression of the log marginal likelihood.

\subsection{Sparse Gaussian Processes}\label{ssec: sparse GP}
One main disadvantage of exact GPs is that the complexity of one prediction step scales quadratically with the dataset size $n_D$~\cite{Hewing2020a}, impacting real-time capabilities, e.g., as a surrogate model for NMPC.
Sparse GPs address this issue by either optimally selecting a subset of the full dataset, or by optimizing for a set of pseudo points $\bm{m} \in \mathbb{R}^{n_m\times d_z} $ that we call ``inducing input''~\cite{Rasmussen2006} as a smaller inference dataset. 
Although this method requires solving a larger optimization problem compared to training an exact GP, the prediction cost is reduced from $O({n_D}^2)$ to $O(n_m^2)$, where $n_m$ is the number of inducing points~\cite{Hewing2020a}.
The selected inference dataset is then used to approximate the true GP posterior distribution, while reducing computational effort.

We implement the pseudo point method, specifically the Variational Free Energy (VFE) approach~\cite{Titsias2009}, wherein the points $\bm{m}$ are chosen to minimize the Kullback-Leibler (KL) divergence between the sparse GP and the true GP posterior, to provide a close sparse GP approximation.

The sparse GP posterior is specified by the mean and covariance, shown in Eq.~\eqref{eq: sparse GP}.
\begin{equation}\label{eq: sparse GP}
    \begin{aligned}
        \mu_f(\bm{z}) &= 
        \sigma_n^{-2}\bm{k}_{\bm{m}}^T
        (\bm{K}_{\bm{mm}} + \sigma_n^{-2}\bm{K}_{\bm{mn}}\bm{K}_{\bm{nm}})^{-1}
        \bm{K}_{\bm{mn}}\bm{Y}\\
        \sigma^2_f(\bm{z}) &=  
        k(\bm{z},\bm{z})- 
        \bm{k}_{\bm{m}}^T\bm{K}_{\bm{mm}}^{-1}\bm{k}_{\bm{m}}\\
        &+\bm{k}_{\bm{m}}^T
        (\bm{K}_{\bm{mm}}+\sigma_n^{-2}\bm{K}_{\bm{mn}}\bm{K}_{\bm{nm}})^{-1}
        \bm{k}_{\bm{m}}
   \end{aligned}
\end{equation}
The scalar $\sigma_n^{-2}$ is the inverse of the noise variance parameter.
Similar to Eq.~\eqref{eq: Gaussian process}, the vector $\bm{k}_{\bm{m}}\in \mathbb{R}^{n_m} $ contains the covariances between the prediction point and $n_m$ inducing inputs.
Analogously, the matrix $\bm{K}_{\bm{mm}} \in \mathbb{R}^{n_m\times n_m} $ contains the covariances between inducing points and the matrix $\bm{K}_{\bm{mn}} \in \mathbb{R}^{n_m\times n_D} $ contains the covariances between inducing points and true data inputs $\bm{Z}$. 

\subsection{State Space Modeling}
We model the dynamic system in a discrete-time nonlinear state space form with the additive measurement noise~\cite{Rawlings2017}:
\begin{equation}\label{eq: nonlinear discrete state space}
    \begin{aligned}
        \bm{x}[j+1] &= \bm{f}(\bm{x}[j], \bm{u}[j]) \\ 
        \bm{y}[j]   &= \bm{x}[j] + \bm{\epsilon}[j] \\
        \bm{y}[j+1] &= \bm{f}(\bm{x}[j], \bm{u}[j]) + \bm{\epsilon}[j]
    \end{aligned}
\end{equation}
Here, $j$ is the discrete time index, $t = j\Delta t $, where $\Delta t$ is a fixed time interval, and $t$ is the real process time.
Furthermore, $\bm{y}[j]\in\mathbb{R}^{d_y}$, $\bm{x}[j]\in\mathbb{R}^{d_x}$, and $\bm{u}[j]\in\mathbb{R}^{d_u}$ are now vectors of system measurements, states, and control inputs at time step $j$, respectively.
We assume all states can be measured, perturbed by i.i.d. Gaussian measurement noise.
The function $\bm{f}(\cdot)$ represents the one-step nonlinear transition function of the time-invariant system, to be approximated using GPs. 

To model the state-space as described in Eq.\eqref{eq: nonlinear discrete state space}, we fit each state output via an independent scalar output GP, which is a common simplification done by assuming independent state predictions~\cite{Bradford2020, Hewing2020a}.
Each output GP takes the full state and control vectors from the previous time step as input. 

We outline the procedure of fitting the function $\bm{f}(\cdot)$ from system trajectories.
From each trajectory of the system, time-series data is presented in the form of $\{\bm{y}[0],\bm{u}[0],...,\bm{y}[T-1],\bm{u}[T-1],\bm{y}[T]\}$ where $T$ is the total number of discrete time steps in each simulation run. 
The training dataset for each output GP is given by:
\begin{equation}\label{eq: training data}
\begin{aligned}
\bm{Y}_i &= 
\begin{bmatrix}
y_i[1],\\
...,\\
y_i[j+1],\\
...,\\
y_i[T]
\end{bmatrix},\quad
\bm{Z}=
\begin{bmatrix}
[\bm{y}[0],\bm{u}[0]],\\
...,\\
[\bm{y}[j],\bm{u}[j]],\\
...,\\
[\bm{y}[T-1],\bm{u}[T-1]]
\end{bmatrix},\\
\forall i &\in \{1,...,d_x\}
\end{aligned}
\end{equation}

Note that in the input matrix $\bm{Z}$, since we do not have access to the true state data $\bm{x}[j]$ for input, the common approach is to substitute with noisy measurements $\bm{y}[j]$ as state inputs, and allowing the output noise variance to be inflated~\cite{Mchutchon2011}.
To meet the GP prior mean assumption, the input and output data are standardized to zero mean and unit variance. 
We fit the sparse GP as outlined in Subsection ~\ref{ssec: sparse GP} with the dataset from Eq.~\eqref{eq: training data}, to obtain $d_x$ output GPs. Hence, the final GPs read as follows:
\begin{equation}\label{eq: GP output model}
    \begin{aligned}
        x_i[j+1] &= f_{\mathcal{GP},i}(
        [\bm{x}[j], \bm{u}[j]]) \\
        f_{\mathcal{GP},i}(
        [\bm{x}[j], \bm{u}[j]])
         &\sim N(\mu_{f,i}(
        [\bm{x}[j], \bm{u}[j]]),
        \sigma^2_{f,i}((
        [\bm{x}[j], \bm{u}[j]]))
    \end{aligned}
\end{equation}

\subsection{Multi-Step Predictions}
For multi-step ahead prediction of the states given a control sequence, we iterate the functions $f_{\mathcal{GP},i}$ in Eq.~\eqref{eq: GP output model} by taking predicted output states as input for predicting further time steps. 
However, as the predicted GP outputs are random variables, this uncertainty needs to be propagated through the nonlinear GP, where inputs are deterministic in the exact case.

Other than Monte-Carlo sampling methods~\cite{Bradford2020}, there have been multiple approximation methods proposed~\cite{Hewing2020a, Girard2002}.
We utilize a moment matching method with a first-order Taylor approximation of the GP mean around prediction points, due to its computational efficiency~\cite{Hewing2020a}.
By propagating via linear transformation, the resulting posterior is Gaussian, and allows us to reiterate for further time steps. 
For the random input 
$\begin{bmatrix}
    \bm{x}[j] \\ \bm{u}[j]
\end{bmatrix} 
$
, the next predicted state can be expressed by their moments as:
\begin{equation}\label{eq: uncertainty propagation}
\begin{aligned}
\mu_i[j+1] &= \mu_{f,i}
([\bm{\mu}_{\bm{x}}[j], \bm{u}[j]])\\
{\sigma^2}_i[j+1] &= \sigma^2_{f,i}([\bm{\mu}_{\bm{x}}[j], \bm{u}[j]]) \\ 
+ \frac{\partial\mu_{f,i}}{\partial\bm{x}}
&([\bm{\mu}_{\bm{x}}[j], \bm{u}[j]])^T\bm{\Sigma}_{\bm{x}}[j]
\frac{\partial\mu_{f,i}}{\partial\bm{x}}([\bm{\mu}_{\bm{x}}[j], \bm{u}[j]])
\end{aligned}
\end{equation}
The functions $\mu_{f,i}$ and ${\sigma^2}_{f,i}$ are the indexed GP functions from Eq.~\eqref{eq: sparse GP} for the predicted state output of $i^{th}$ dimension, $x_i[j+1]$.
The vector $\bm{\mu}_{\bm{x}}[j]$ consists of the mean value of each state dimension as input. 
By assuming independent prediction states, the covariance matrix between the inputs is a diagonal matrix of the state variances, i.e., ${\bm{\Sigma}_{\bm{x}}}_{ii}= {\sigma^2}_i^x$.
The term $\frac{\partial\mu_i}{\partial\bm{x}}$ is the partial derivative of the GP mean function with respect to its state inputs.

\section{Stochastic (economic) GP-based nonlinear model predictive control}\label{sec: GP-NMPC}
\subsection{Gaussian process-based nonlinear model predictive control (GP-NMPC)}
We formulate the stochastic optimal control problem (OCP) for the GP-NMPC controller at sampling instant $l$:
\begin{equation}\label{eq: OCP}
    \begin{aligned}
    \underset{\hat{\bm{u}}}{\text{min}}\ \quad \sum_{j=0}^{N_h-1}\mathbb{E}[l(&\hat{\bm{x}}[j],\hat{\bm{u}}[j])] + \beta\mathbb{E}[l_f(\hat{\bm{x}}[N_h])]\\ 
    s.t.\quad 
    \hat{x}_i[j+1] &\sim N(\mu_i^x[j+1],\sigma^2_i[j+1] )\\
    \mu_i[j+1] &= \mu_{f,i}
([\bm{\mu}_{\bm{x}}[j], \hat{\bm{u}}[j]])\\
{\sigma^2}_i[j+1] &= \sigma^2_{f,i}([\bm{\mu}_{\bm{x}}[j], \hat{\bm{u}}[j]]) \\ 
+ \frac{\partial\mu_{f,i}}{\partial\bm{x}}&
([\bm{\mu}_{\bm{x}}[j], \hat{\bm{u}}[j]])^T\bm{\Sigma}_{\bm{x}}[j]
\frac{\partial\mu_{f,i}}{\partial\bm{x}}([\bm{\mu}_{\bm{x}}[j], \hat{\bm{u}}[j]])\\
    \mu_i[0] = y_i[l]&,\quad \sigma^2_i[0] = \sigma_{n,i}^2\\
    \hat{\bm{x}}[j] &= \big[\hat{x}_1[j],...,\hat{x}_{d_x}[j]\big]^T\\
     i \in [1,d_x],&\quad j \in [0,N_h-1] 
    \end{aligned}
\end{equation} 

The NMPC variables $\hat{\bm{x}}$, $\hat{\bm{u}}$ represent the NMPC states and control input computed for the prediction horizon $N_h$. 
The expected value of the objective function $l(\cdot,\cdot)$ and terminal objective $l_f(\cdot)$ over the probabilistic state predictions is minimized in this formulation.
The terminal objective is weighted by the constant $\beta$.

The mean $\mu_i[j]$ and variance $\sigma^2_i[j]$ of each independent state are predicted by their corresponding GP functions from Eq.~\eqref{eq: uncertainty propagation}.
The GP functions take the full state vector and control vectors as input, as outlined in Section~\ref{sec: GP_state_space}.

As we have no access to the true state values of the system at each sampling instant, we approximate the initial state $\hat{x}[0]$ as a normal distribution for every OCP during the control loop, since Gaussian noise is assumed.
This state distribution has a mean at the deterministic measurements $y_i[l]$.
Since the noise variance hyperparameter $\sigma^2_{n,i}$ from GP training provides an approximation of the true system noise variance $\sigma^2_\epsilon$, they are used as variance for the initial state distribution $\sigma^2_i[0]$.

The OCP in Eq.~\eqref{eq: OCP} is solved for the prediction horizon at each sampling instant $l$, with new feedback state values $x[l]$ as the system moves forward in time.
This online optimization results in the optimal control profile, $\hat{\bm{u}}^*$.
The moving horizon implementation gives the MPC control law as $\bm{u}[l] = \hat{\bm{u}}^*[j=0] $, where the first element of the optimal control sequence, $\hat{\bm{u}}^*$, is applied to the system.

\subsection{Objective functions}\label{ssec: objective functions}
Given that the predicted states are Gaussian variables, we minimize the expected value of the objective functions $l()$ over the state distributions. 
As a simplification, we look at common quadratic and linear functions, which cover most general control objectives. 
Specifically, we investigate a quadratic tracking objective and a linear economic objective.

For the common setpoint-tracking task with quadratic cost, the expected value can be reformulated in terms of the mean and variance of states with distributed values~\cite{Paulson2020}. 
For the tracking of a single state $\hat{\bm{x}}_i$ at setpoint $x_{set}$, the objective function is written as:
\begin{equation*}
\begin{aligned} 
    \mathbb{E}[l_{tracking}(\hat{\bm{x}}[j],\hat{\bm{u}}[j])]&=
    \mathbb{E}
    [(\hat{x}_i[j]-x_{set})^2]\\
    &=(\mu_i[j]-x_{set})^2 + \sigma^2_i[j]
\end{aligned}
\end{equation*}

The terminal objective function $l_f()$ is specified to be equivalent to the stage cost function $l()$.
Note that the trace of the state covariance matrix appears in this reformulation of the quadratic form of a random variable.
This minimization of the expected squared deviation penalizes control actions that lead to high predicted state variances.
Therefore, the objective favors control sequences that result in high confidence predictions.

We investigate the formulation of an economic objective, e.g., the sum of product mass over duration.
We maximize the expected value of the objective function, where $\hat{x}_{econ}$ is the state to be maximized in Eq.~\eqref{eq: economic obj}. The objective to be maximized is simply the sum over the state mean vector.
\begin{equation*}\label{eq: economic obj}
\begin{aligned}
    \mathbb{E}[l_{econ}(\hat{\bm{x}}[j],\hat{\bm{u}}[j])]&=
    \mathbb{E}
    \left[\hat{x}_{econ}[j]\right]\\
    &=\mu_{prod}[j]
\end{aligned}
\end{equation*}

\subsection{Chance Constraints for Safe Control under Uncertainty}
A significant advantage of GP models is that they provide prediction uncertainty based on observed data.
The uncertainty quantification allows the controller to identify high and low confidence regions of the prediction model.
By implementing chance constraints ~\cite{Hewing2020a, Paulson2020}, the probability mass of constraint violations is restricted, i.e., the optimizer only selects control actions that fulfill the specified confidence of safety.
Thus, the chance constraints enforce cautious behavior during learning and operation, and address safety issues such as over-extrapolation.
These constraints are especially important when the GP model has limited observations of the system and is susceptible to plant-model mismatch.

Even though the probability distribution for nonlinear constraint violations is generally intractable and can only be evaluated with Monte-Carlo methods, we consider simple linear inequality constraints, which are common in applications. 
As described in \cite{Hewing2020a}, probabilistic formulations of linear inequality constraints $\bm{h}^T\hat{\bm{x}}[j] \leq b $ can be reformulated in terms of the moments as such:
\begin{equation}\label{eq: chance constraints}
\begin{aligned}
    Pr(\bm{h}^T\hat{\bm{x}}[j] \leq b) &\geq \epsilon\\
    \bm{h}^T\hat{\bm{\mu}}_x[j] &\leq b - \phi^{-1}(\epsilon)\sqrt{\bm{h}^T\Sigma_x[j]\bm{h}}\\
    \Sigma_x[j] &= diag(\bm{\sigma}^2_x[j]) 
\end{aligned}
\end{equation}

The constant $\epsilon$ represents the desired minimum probability of constraint violation at each time step. 
Function $\phi^{-1}()$ is the inverse cumulative distribution function (CDF) of a standard Gaussian. 
This reformulation in the explicit form allows fast online evaluations of these chance constraints. 

It occurs that these constraints can be over-restrictive given a sparse dataset and tight bounds on certain states. 
These tight constraints lead to an infeasible solution space of the OCP, and no resulting output from the NMPC controller.
Therefore, we implement selected chance constraints~\eqref{eq: chance constraints} as soft constraints in the objective:
\begin{equation} \label{eq: soft constraints}
    \begin{aligned}
         \underset{\hat{\bm{u}},s}{\text{min}}\ \quad \sum_{j=0}^{N_h-1}\mathbb{E}[l(&\hat{\bm{x}}[j],\hat{\bm{u}}[j])] + \mathbb{E}[l_f(\hat{\bm{x}}[N_h],\hat{\bm{u}}[N_h])] + \alpha s \\
         s.t.\quad s \geq\ &\bm{h}^T\hat{\bm{\mu}}_x[j] - b + \phi^{-1}(1-\epsilon)\sqrt{\bm{h}^T\Sigma_x[j]\bm{h}}\\
         s\geq \ &0.
    \end{aligned}
\end{equation}
The soft constraints are implemented via the addition of slack variables $s$, which avoids infeasible solution spaces for the optimization problem. 
The coefficient $\alpha$ represents the weight for the soft constraint in the objective.
With soft constraints, we can no longer guarantee the satisfaction of selected constraints to the desired confidence, but the controller is still heavily penalized for unsafe actions, hence minimizing the risk of constraint violation.

\section{Model Learning GP-NMPC for (Semi-)Batch Processes}\label{sec: learning scheme}
In this section, we present our concept of the GP-MLMPC for (semi-)batch processes. The proposed model learning scheme alternates between updating the GP state space model and applying the GP-NMPC controller over a batch duration. 
\begin{figure*}
    \centering
    \includegraphics[width=1\linewidth]{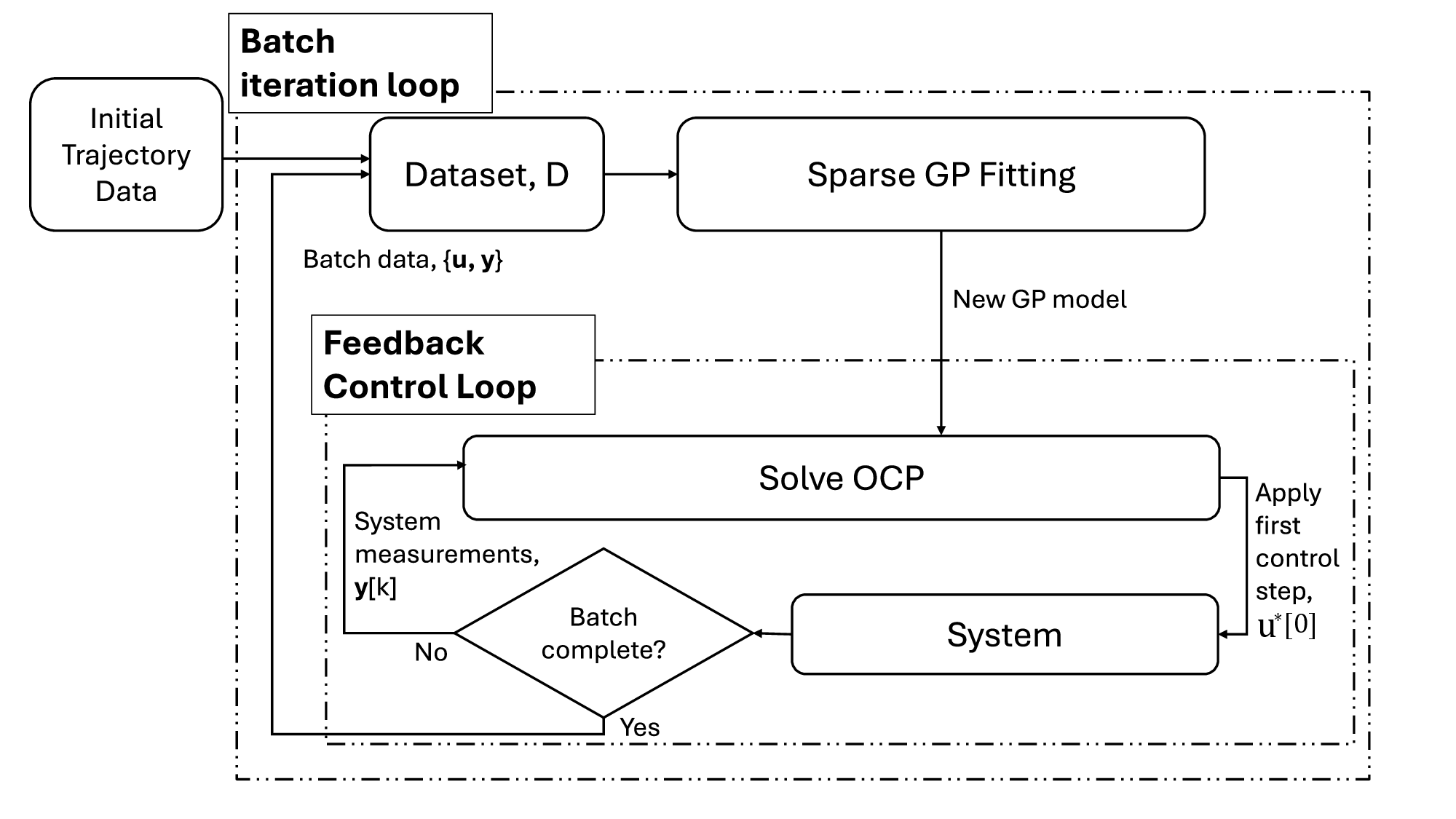}
    \caption{Batch iteration learning scheme via GP-MLMPC. The inner loop represents feedback control within a batch duration.}
    \label{fig: iterative learning}
\end{figure*} 
As illustrated in Fig.~\ref{fig: iterative learning}, the scheme consists of two loops, where the outer loop performs batch-to-batch updates of the GP model to progressively improve control performance, while the inner loop corresponds to the temporal feedback control loop.
After each batch iteration $B\in \{1,...,N_{batch}\}$, the training dataset $\mathcal{D}_B$, is augmented with measurement data collected during the previous batch, i.e.,
\begin{equation}
    \mathcal{D}_{B} =\bigcup_{b = 0}^B  \mathcal{D}_b 
\end{equation}
We refit the sparse GP state space model in Eq.~\eqref{eq: GP output model} with $\mathcal{D}_B$ and obtain the updated GP parameters.
In the feedback control loop, we apply the updated GP-NMPC to the system in a moving-horizon fashion, as described in Section~\ref{sec: GP-NMPC}.
The inner loop is completed once the fixed batch duration has been reached and the batch is completed.

The system measurements collected during the completed batch are passed to the next batch iteration loop, for updating the dataset $\mathcal{D}_{B+1}$.
The updated dataset encompasses all measurements collected up to and including batch iteration $B$.
Using this dataset, we re-optimize the GP hyperparameters and inducing points to update the GP state space model for the controller, as described in Section~\ref{sec: GP_state_space}.
The re-optimization ensures that the resulting GP distribution provides the most probable representation of the underlying system dynamics, given the available data.
Particularly during the early batch iterations of the learning scheme, when the dataset is small, re-optimizing the hyperparameters is crucial to capture significant changes as new data becomes available.

With each batch iteration loop, we hypothesize that the learning-based GP-NMPC controller efficiently explores the operating space.
Starting from a sub-optimal trajectory, the controller utilizes high-confidence predictions from the GP model, which results in subsequent improvements, and more importantly, exploration towards the optimal trajectory.
Therefore, the proposed approach allows for iterative learning of uncertain systems via GP-NMPC, starting from a suboptimal initial trajectory.

\subsection{Initial dataset generation}
To initiate the batch iteration loop, we collect observations from a single system trajectory generated using a tuned Proportional-Integral (PI) controller. 
From the collected measurements, we construct the initial dataset $\mathcal{D}_0$, as defined in Eq.~\eqref{eq: training data}.  
This setup reflects a practical scenario in which suboptimal control is applied using a conventional PI controller due to limited knowledge of the system dynamics.
In practice, any available dataset of the system may be used for initialization, e.g., data obtained from manual operation by a technician. 

In related work, surrogate models such as GPs or other machine learning models are often trained to accurately represent the system dynamics over a wide region of the state-control space.
This is commonly achieved through homogeneous sampling of high-fidelity simulations~\cite{Bradford2020} or the design of excitation control signals~\cite{Hewing2020b}.
As a result, a much larger and more informative initial dataset can be obtained to fit the surrogate model to the system.
However, high-fidelity models and highly exciting control signals are often not applicable to real industrial systems, hence limiting the deployment of learning MPC.
In contrast, to mimic the challenge of small data in real industrial settings, we utilize only a single trajectory for initialization.
Thus, the quantity of data and its coverage over the operating space are deliberately limited.
The collected measurements lie sparsely across the state space and are determined by the applied controller and system dynamics. 

\section{GP-MLMPC on Semi-batch Polymerization Reactor}\label{sec: case study and results}
We apply the proposed GP-MLMPC to a semi-batch polymerization reactor case study and show the improvement over batch iterations with the proposed learning loop.
We discuss the results for two different control objectives, namely setpoint-tracking and economic. 

\subsection{Semi-batch Polymerization Reactor}
\begin{figure}
    \centering
    \includegraphics[width=1\linewidth]{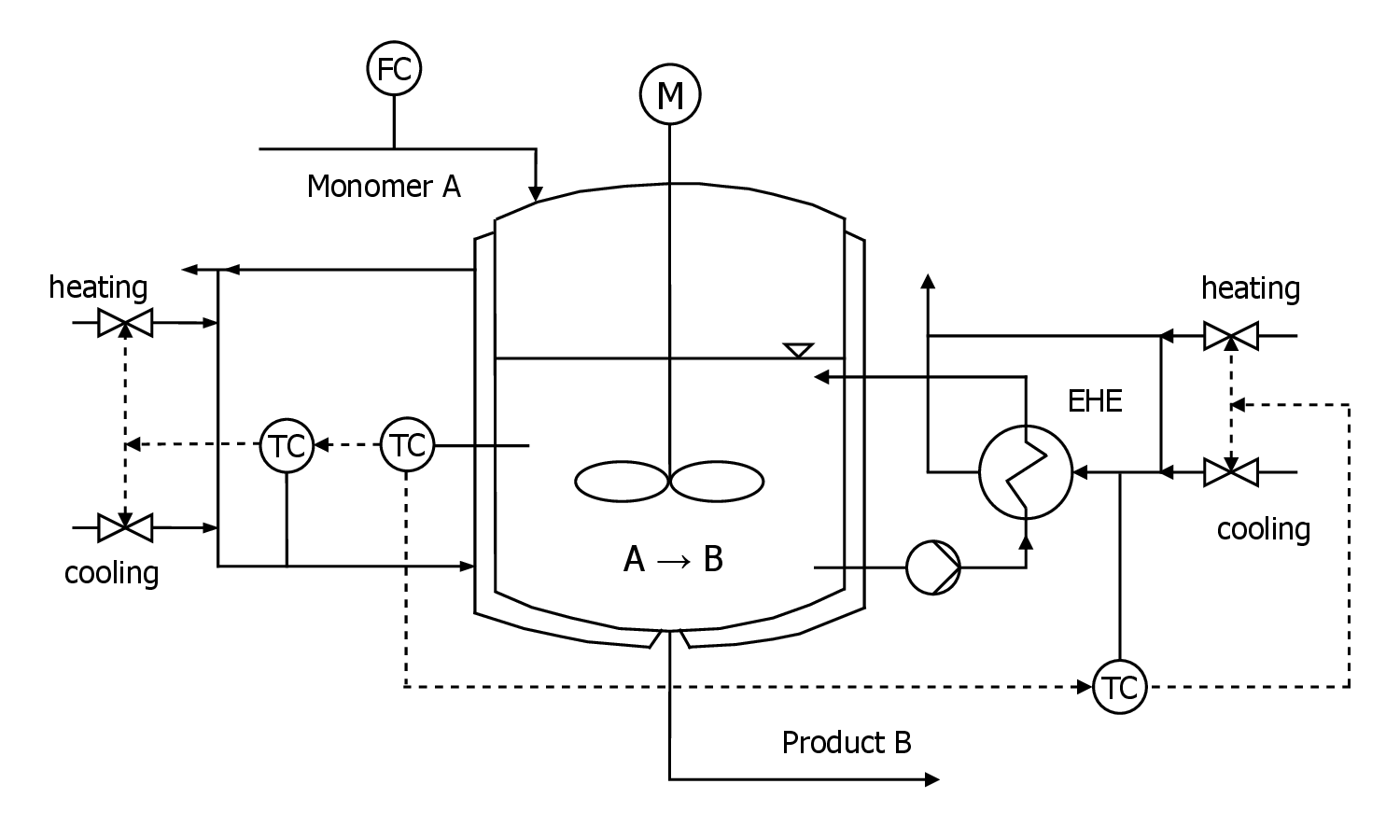}
    \caption{Schematic of semi-batch polymerization reactor with external heat exchanger and cooling jacket, adapted from \cite{Lucia2014}.}
    \label{fig:semi-batch reactor}
\end{figure}

For the control of polymerization reactors, MPC of different varieties has been widely studied~\cite{Silva2008, Joy2019, Faust2021}. 
The reactor case study used in this paper was modeled and investigated by Lucia~et~al.~\cite{Lucia2014} and Rostampour~et~al.~\cite{Rostampour2015}, both for the implementation of NMPC. 
It consists of a semi-batch reactor, into which monomer is fed for reaction, illustrated in Fig.~\ref{fig:semi-batch reactor}. 
The highly exothermic polymerization reaction is regulated via two cooling loops, the reactor cooling jacket and an external heat exchanger. 
Tight temperature control is crucial, both for safety and for the consistent quality of the resulting product. 

The system consists of 10 states and 3 control variables, and measurements are obtained by adding noise to all states. 
The states are mass of water, monomer, and polymer in the system (respectively $m_W$, $m_A$, $m_P$) and temperature of reactor, vessel, jacket, and heat exchanger cold and hot sides (respectively $T_R$, $T_S$, $T_M$, $T_{EK}$, $T_{AWT}$). 
The control variables refer to the feed inlet flowrate  $\dot{m}_{feed}$, jacket inlet temperature $T_{M,in}$, and heat exchanger cold side inlet $T_{AWR,in}$.
For detailed modeling equations and parameters, please refer to Appendix B and the original publication by Lucia~et~al.~\cite{Lucia2014}.

The process noise $\epsilon [j]$ for measurements $\bm{y}[j]$ is sampled from normal distributions. 
We determine nominal standard deviation values from common equipment specifications for temperature measurements, $\sigma_{\epsilon,temp} = 0.1^\circ\text{C}$~\cite{TempAccuracy}, and mass measurements, $\sigma_{\epsilon,mass} = 33 \text{kg}$~\cite{MassAccuracy}.  

We study the tracking and economic objectives in Subsection~\ref{ssec: objective functions}.
For the tracking objective, the reactor temperature $T_R$ is to be maintained at $T_{set} = 90^\circ\text{C}$, i.e., $l_{tracking} = (T_R-T_{set})^2$.
The economic objective involves maximization of polymer mass $m_P$ over the prediction horizon, i.e., $l_{econ} = m_P$.

Both tracking and economic objectives are investigated by simulating the process in \textit{silico}, and are discussed in subsequent sections.
The process is operated under a couple of important constraints.
First, the reactor temperature is required to remain within a tight band of $\pm 2^\circ\text{C}$ around the desired reaction temperature $T_{set} = 90^\circ\text{C}$.
This constraint should be satisfied to reach the desired properties of the product~\cite{Lucia2014}.
In addition, the adiabatic temperature, $T_{adiab}$, which is defined as the maximum temperature the reactor could reach under a complete cooling failure, is constrained to ensure process safety.
The temperature value of $T_{adiab}$ should be maintained below $109^\circ\text{C}$ throughout operation.
The total duration of the reaction is fixed at 2 hours. 

\begin{figure}
    \centering
    \includegraphics[width=1\linewidth]{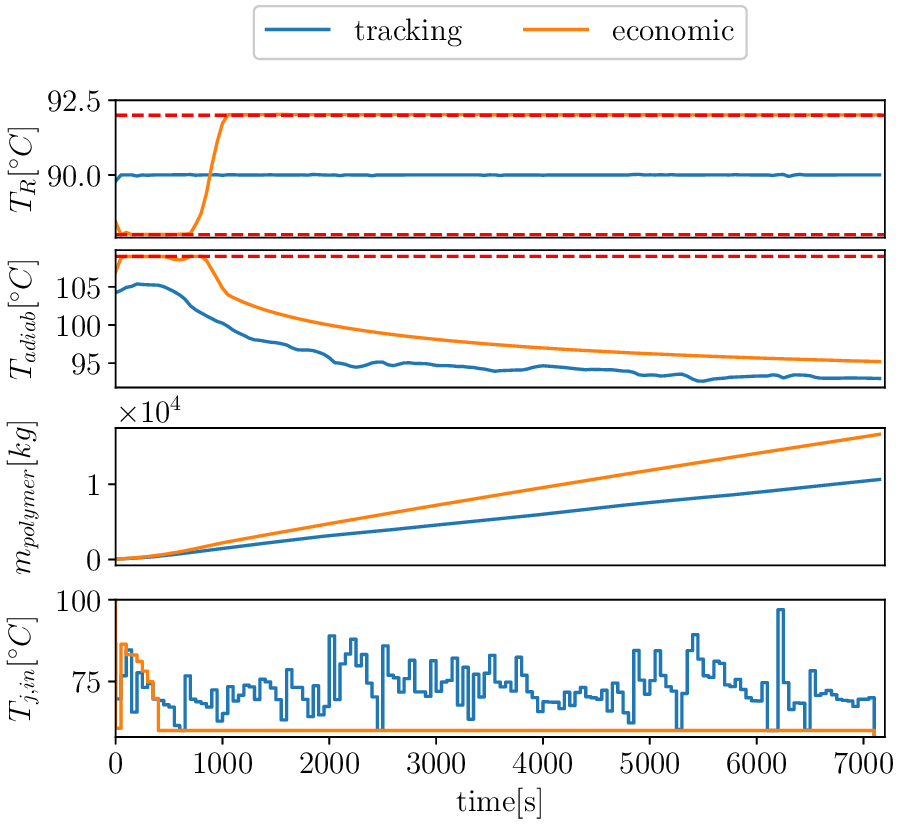}
    \caption{Reactor temperature, adiabatic temperature, mass of polymer, and jacket inlet temperature for tracking and economic NMPC using the perfect model without noise. Constraints are shown in red.}
    \label{fig:solution}
\end{figure}

As a reference for later results, Fig.~\ref{fig:solution} shows the open-loop control solution from dynamic optimization of the full model, where results from both tracking and economic objectives are shown.
Both optimal trajectories serve as a benchmark for GP-MLMPC results in later sections.
The system is modeled and simulated in Pyomo.DAE~\cite{Nicholson2018} and solved with IPOPT~\cite{Wachter2006}.
For the tracking task, a very steady reaction temperature can be achieved in the optimal case.
We observe significantly different control regimes between the two objectives, especially when the polymer mass stays constant for the tracking task to avoid any reaction heat. 
For the economic task, we observe that the optimal reactor temperature trajectory stays on the upper constraint for most of the batch duration.
This trend in reactor temperature illustrates that the economic objectives tend to push systems towards the constraint bounds.

\subsection{Closed-Loop Tracking GP-MLMPC} \label{ssec: tracking results}
We apply the GP-MLMPC scheme on the polymerization case study for $N_{batch} = 10$ batch iterations to observe improvement in control performance.
We run 20 independent simulations to demonstrate variance from measurement noise, and this is shown in the following figures. 

In this implementation, the NMPC sampling interval is set at $\Delta t = 50 \text{s}$, with both prediction and control horizons defined as $N_h = 12$ steps.
The GP state space model is set up with a zero mean prior, while the package GPy~\cite{gpy2014} is used for training the sparse GP models.
We set the number of inducing points for the sparse GP to $m = 20$. 
The NMPC problem is solved with the optimization tool CasADi~\cite{Andersson2019} with IPOPT~\cite{Wachter2006} as solver.

To provide the starting dataset, we implement the PI control on the system for tracking.
The PID control with derivative term leads to large oscillations in states and controls, due to the high noise variance. 
Therefore, we implement PI control for this system.
Using the Ziegler-Nichols tuning rule~\cite{Ziegler1942}, we tune the PI controller with the following settings, $k_P = 114, k_I = 0.3$, to perform the tracking task at $T_{set}$.
The PI controller is implemented to manipulate the cooling jacket inlet temperature, $T_{j,in}$.
The cooling jacket inlet temperature is chosen out of the three available control variables due to its linear response and low delay.
Although the PI controller can be extended to multiple input systems via control allocation or decoupling, the single-input controller here is sufficient to provide a starting dataset and benchmark as a suboptimal trajectory.
The other control variables are left with nominal constant profiles, where $\dot{m}_{feed} = 5000\text{kg h}^{-1},T_{AWT,in} = 70^\circ\text{C}$.

\begin{figure*}
    \begin{subfigure}{0.49\textwidth}
        \centering
        \includegraphics[width=1\linewidth]{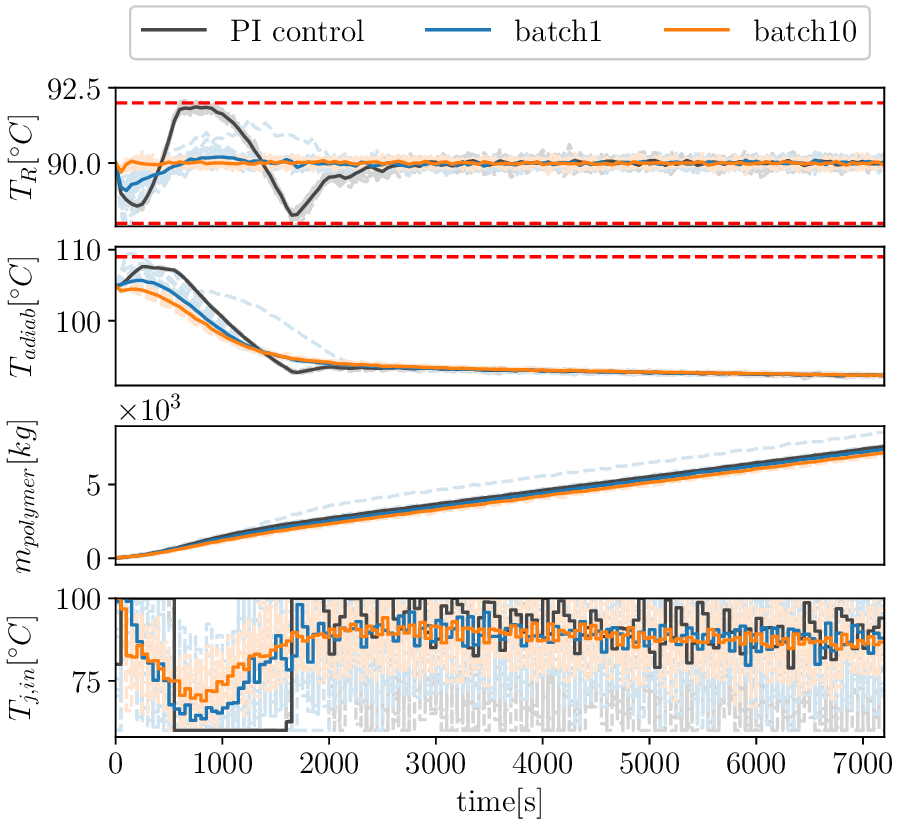}
        \caption{No chance constraint.}
        \label{fig:tracking_no_cc}
    \end{subfigure}
    \begin{subfigure}{0.49\textwidth}
        \centering
        \includegraphics[width=1\linewidth]{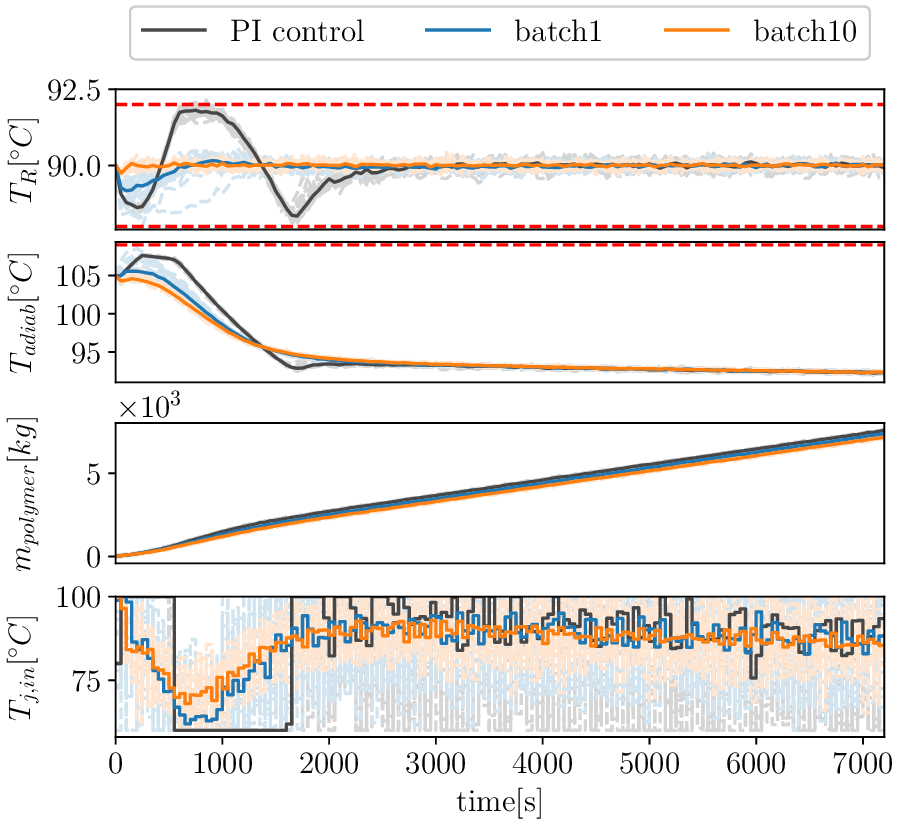}
        \caption{With chance constraint.}
        \label{fig:tracking_2_cc}
    \end{subfigure}
    \caption{Reactor temperature, adiabatic temperature, mass of polymer, and jacket inlet temperature from tracking GP-NMPC without chance constraints (\subref{fig:tracking_no_cc}) and with chance constraints (\subref{fig:tracking_2_cc}).
    Batch Iterations $B=1$ and $B=10$ are shown, with PI control as the benchmark.
    Each light dashed line represents a simulation run, and the solid line represents the median value. Constraints are shown in red.}
    \label{fig:EE}
\end{figure*}

Fig.~\ref{fig:EE} shows the closed-loop state and control trajectories of the tracking GP-NMPC, which compares batch iteration 1 and 10 to demonstrate the learning capabilities of the approach.
We also plot here the PI trajectory as a benchmark to observe the difference in performance.
We also compare the effectiveness of chance constraints, where in the left plots only the mean state trajectory is constrained (no chance constraint), while in the right plots chance constraints are applied.
We implemented a chance constraint probability of $\epsilon = 95\%$, alongside a soft constraint coefficient of $\alpha =1000$, as described in Eq.~\eqref{eq: soft constraints}.
From Fig.~\ref{fig:EE}, both plots show a significant improvement in tracking performance in comparison to the PI controller. 
Already for $B=1$, we obtain much less deviation when observing the start of the batch process, where an overshoot in $T_R$ is clearly present from the PI controller.
The more conservative action is attributed to the predictive capabilities of the controller.
The controller also manipulates multiple control variables to steer the system, utilizing all degrees of freedom.
By $B=10$, nearly no deviation can be observed from the setpoint, and we obtain near-optimal performance compared to Fig.~\ref{fig:solution}.
Hence, the GP-MLMPC scheme demonstrates the capability to learn the nonlinear system dynamics starting from a single suboptimal trajectory. 

Notably, there is no observable difference between applying and not applying chance constraints in this scenario.
Both plots in Fig.~\ref{fig:EE} have no constraint violations.
This lack of violation likely stems from inherent variance minimization through the stochastic quadratic objective as discussed in Subsection~\ref{ssec: objective functions}. 
In this scenario, the reactor temperature setpoint is positioned at the midpoint of the two inequality constraints.
Thus, the controller steers the system towards a safe operating region far from constraints.
Thus, chance constraints are not needed in this scenario; still, it is practically important that they are present to provide guarantees for process safety up to a high confidence level.

\begin{figure}
    \centering
    \includegraphics[width=1\linewidth]{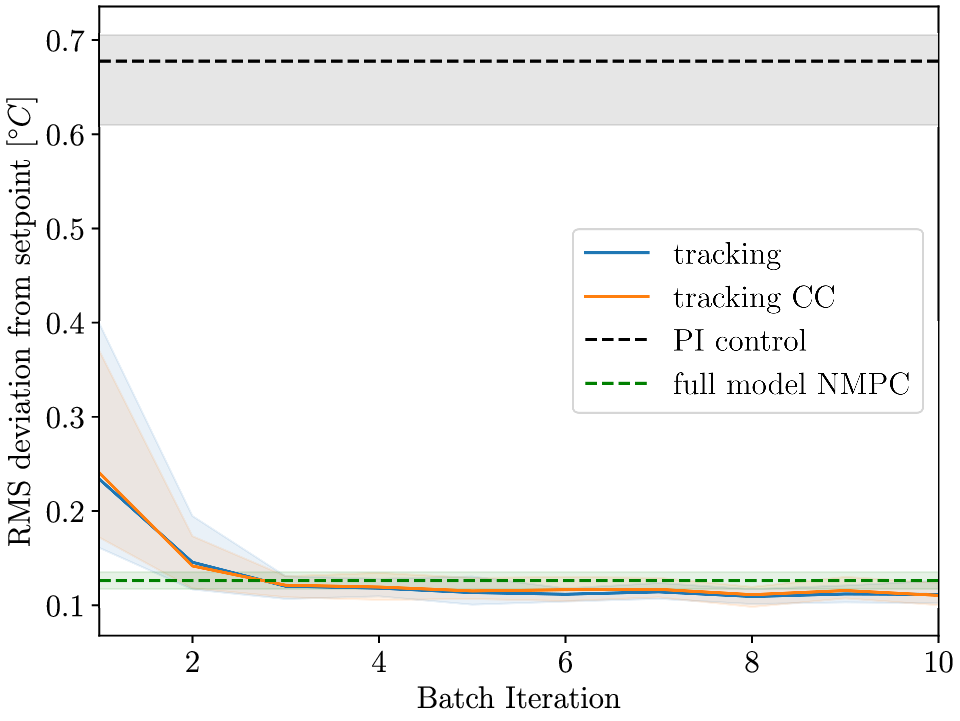}
    \caption{Root mean square deviation from setpoint of tracking MPC decreases over batch iterations to achieve on-par performance with full-model NMPC. PI and full model NMPC are shown as benchmarks. The shaded Region represents the 5th -95th percentile, and the solid line represents the median values.}
    \label{fig:tracking_error}
\end{figure}

In Fig.~\ref{fig:tracking_error}, the root-mean-squared deviation from $T_{R,set}$ for each batch iteration is plotted.
The solid line represents median values over 20 simulations, while the shaded region represents the 5th-95th percentile.
Fig.~\ref{fig:tracking_error} confirms an overall decreasing trend for both GP-NMPC controllers, where it converges by around $B=3$.
The shrinking of the shaded region further supports this convergence.

The results show that the GP-MLMPC achieved significant improvement when compared to the PI control benchmark.
In both GP-MLMPC, with and without chance constraints, the tracking performance converges closely with that of the full-model NMPC benchmark.
Interestingly, in later iterations, the error from GP-MLMPC is lower than the full-model NMPC benchmark.
Here, the variance minimization of the GP-NMPC controller offers a plausible explanation. 
As the controller selects control actions that lead to less uncertainty, aggressive actions that overcompensate are avoided as they lead to regions of high uncertainty.
Instead, more conservative actions are chosen, leading to a slightly steadier performance overall.
On the other hand, the full-model NMPC is susceptible to measurement noise in the system feedback loop.
These noise signals trigger responses from the full-model NMPC to steer back the system, even though there is no actual deviation in the simulated system.

The fast convergence in performance can be attributed to the data efficiency of GPs as a non-parametric model.
The result agrees with the literature where GPs have been shown to perform well in modeling accuracy with limited data, as their flexibility is adapted to the sample size~\cite{Myren2021}.
Furthermore, the RBF kernel has a strong prior assumption of smooth functions, which acts as a strong inductive bias~\cite{Micchelli2006}.
This performs well if the underlying kernel assumption matches the system dynamics, as it is for this case study, where only smooth dynamics are present. 
In comparison, although neural networks are very flexible, they are also more prone to overfitting, especially when a large amount of weights are trained on small data~\cite{Goldin2023}.
Therefore, the data-efficient property of Gaussian Processes is advantageous for learning from limited operational measurements in real systems.

With the learning scheme, the GPs are updated with new observations every batch iteration along the operating region near the setpoint.
Thus, the fitted GPs for higher iterations show higher accuracy near the space of optimal decisions and state trajectories.
This, in turn, supports optimal control decisions for later iterations. 
Although the fitted GPs do not generalize to accurate surrogate models over the whole domain, this selective learning around optimal trajectories presents a \emph{data-efficient} approach to achieve an equal level of performance as shown by the full model NMPC.

\subsection{Closed-Loop Economic GP-MLMPC} \label{ssec: economic results}

\begin{figure*}
    \begin{subfigure}{0.49\textwidth}
        \centering
        \includegraphics[width=1\linewidth]{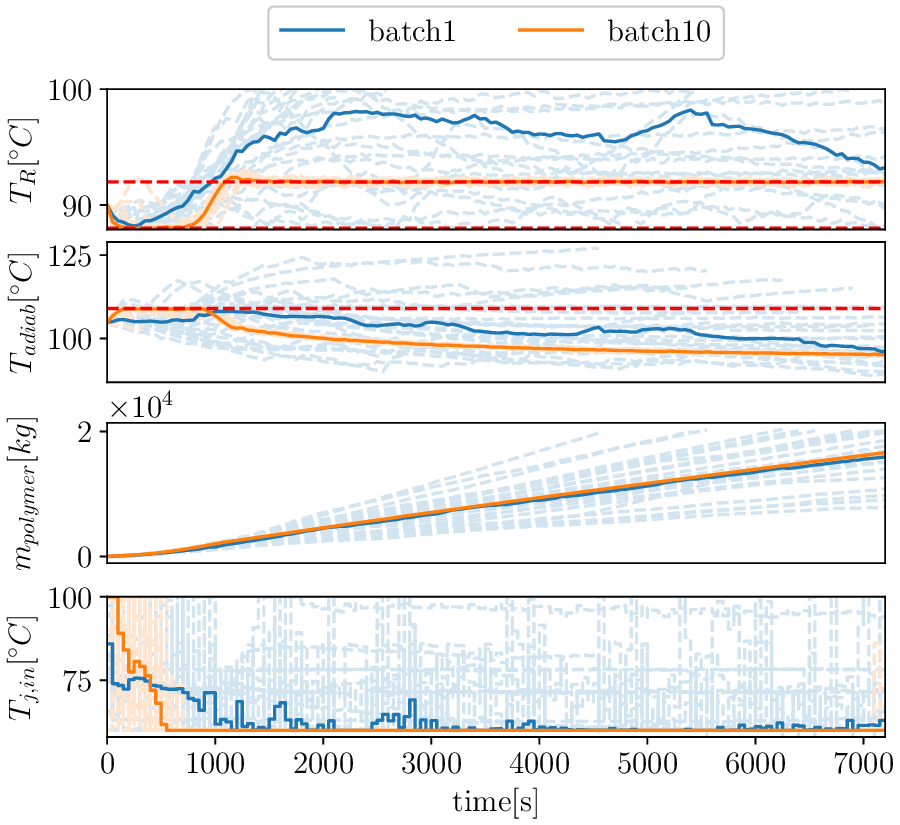}
        \caption{}
        \label{fig:economic_no_cc}
    \end{subfigure}
    \begin{subfigure}{0.49\textwidth}
        \centering
        \includegraphics[width=1\linewidth]{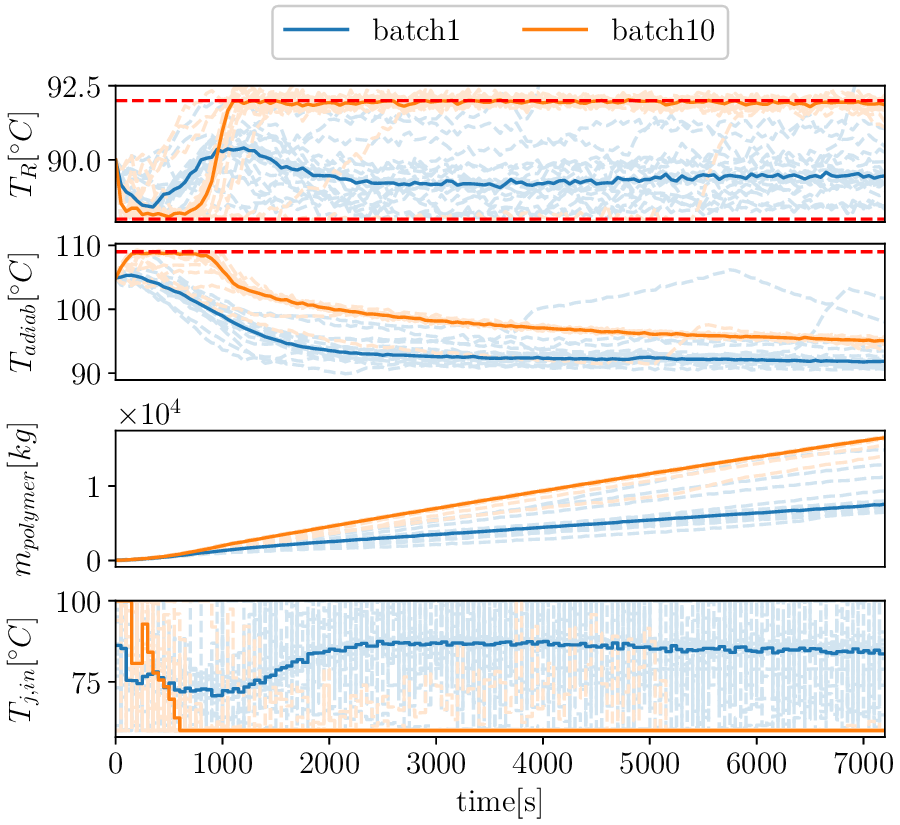}
        \caption{}
        \label{fig:economic_2_cc}
    \end{subfigure}
    \caption{Reactor temperature, adiabatic temperature, mass of polymer, and jacket inlet temperature from economic GP-NMPC without chance constraints (\subref{fig:economic_no_cc}) and with chance constraints (\subref{fig:economic_2_cc}). 
    Batch Iterations $B=1$ and $B=10$ are shown.
    Each light dashed line represents a simulation run, and the solid line represents the median values. 
    Constraints are shown in red.}
    \label{fig:MP}
\end{figure*}

We show results from the economic task under the same NMPC settings as the tracking task.
Fig.~\ref{fig:MP} shows the closed-loop trajectories for the GP-MLMPC, for batch iteration $B = 1$ and $B=10$.
Similarly to the plots in Subsection ~\ref{ssec: tracking results}, in the left plots only the state means are constrained, while in the right plots chance constraints are applied.

In the economic case, the chance constraints do play a significant role in the constraint satisfaction of the system.
In the case without chance constraints (Fig.~\ref{fig:economic_no_cc}), severe violations in both quality and safety constraints for $B=1$, when only mean predictions of the state, $\mu_x$, are constrained.
This setup mimics cases where surrogate models without uncertainty quantification are used.
So, the prediction uncertainty is not accounted for by the objective function or the constraints. 
Therefore, the controller is allowed to choose actions that lead to high prediction uncertainty, even though these actions are often extrapolating from data and have a high probability of leading to unsafe regions.
Such unsafe learning and operation is unrealistic and not applicable for the control of chemical processes.

On the contrary, under $95\%$ chance constraint in Fig.~\ref{fig:economic_2_cc}, we observe a more conservative and consistent performance on $B=1$.
We observe no constraint violation in $B=1$ across 20 simulations.  
With the presence of chance constraints, we see that in the first batch iteration, the performance is actually similar to that of the tracking objectives. 
The trajectory shows that the controller chooses to operate within a high-confidence state-action space, in this case, where PI data are most concentrated.
The comparison clearly indicates the advantage of uncertainty quantification by the probabilistic GPs to operate in unknown and uncertain systems.

\begin{figure*}\begin{subfigure}{0.49\textwidth}
        \centering
        \includegraphics[width=1\linewidth]{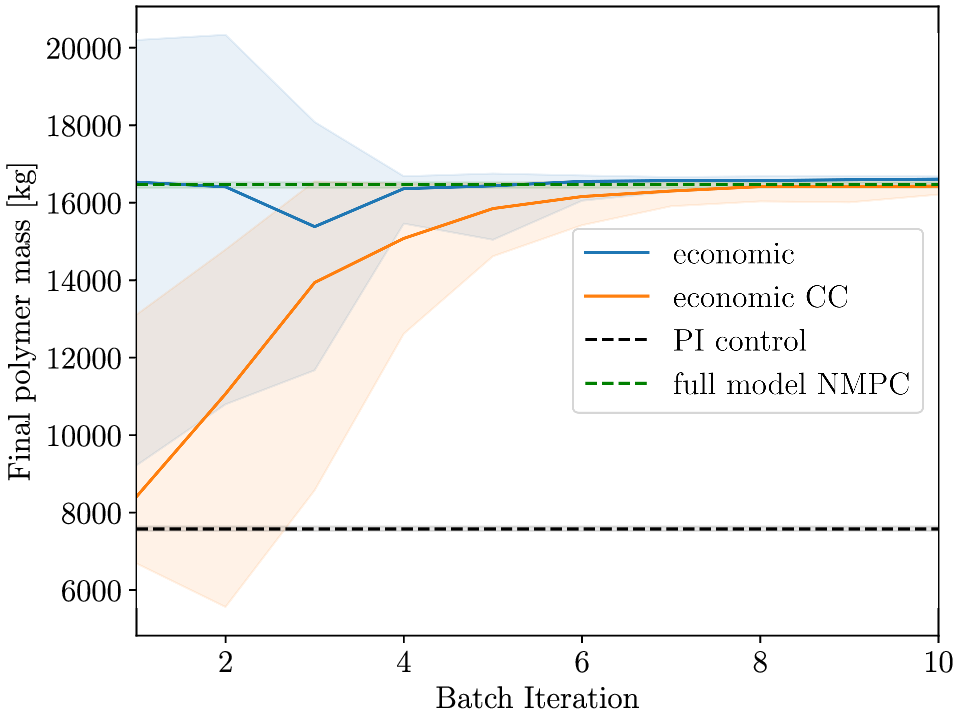}
        \caption{}
        \label{fig:product_mass}
    \end{subfigure}
    \begin{subfigure}{0.49\textwidth}
        \centering
        \includegraphics[width=1\linewidth]{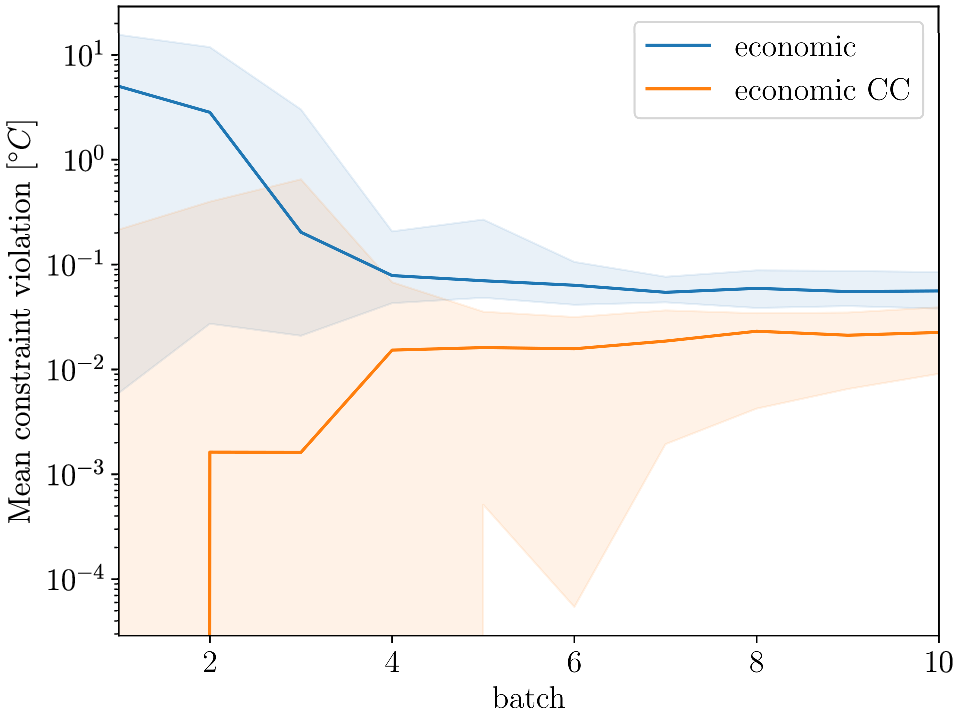}
        \caption{}
        \label{fig:constraint_violation}
    \end{subfigure}
    \caption{(\subref{fig:product_mass}) Final polymer mass at end of batch duration of economic GP-MLMPC. (\subref{fig:constraint_violation}) Mean constraint violation over the full batch duration of the system, comparing the effect of chance constraints (CC). PI and full model NMPC are shown as benchmarks. The shaded Region represents the 5th -95th percentile, and the solid line represents the mean values.}
    \label{fig:improvement}
\end{figure*}

In Fig.~\ref{fig:product_mass}, we observe the batch-wise improvement by the GP-MLMPC scheme in final product mass, with the PI control and full model NMPC as benchmark.
Fig.~\ref{fig:constraint_violation} shows the mean constraint violation across discrete time steps in full batch duration, for the GP-MLMPC schemes.
The solid lines represent the mean value of 20 runs, while the shaded region shows the 5th to 95th percentile.

The Fig.~\ref{fig:product_mass} shows that GP-MLMPC without chance constraints has a higher final product mass for all batch iterations, even at the starting iteration $B=1$.
However, Fig.~\ref{fig:constraint_violation} shows the complete picture.
We observe that without chance constraints, significant constraint violations occur for the first few batches, in line with Fig.~\ref{fig:economic_no_cc}.
Even though the task of achieving a high final yield is achieved, the added difficulty lies in satisfying the imposed constraints.
Without taking into account the prediction confidence, the controller frequently steers the system towards uncertain operating regions.
In these uncertain regions, the model frequently extrapolates, and the mean constraints are insufficient in maintaining safe control.
On the other hand, the same controller with chance constraints (in orange) started with much lower product mass, and steadily increases over the batch iterations.
Although slower improvement is seen, we can observe that the constraint violations are 1-2 magnitudes lower than those of the GP-NMPC without chance constraints. 
The introduction of chance constraints is advantageous in guiding the GP-NMPC controller in the safe operation of unknown systems.

Note that, in the event of constraint violations, we allow the simulation of the batch process to proceed and collect trajectory data for further iterations. 
However, this setup is an unrealistic scenario that we use for comparison.
In the absence of chance constraints, Fig.~\ref{fig:constraint_violation} indicates that the controller still allows for large violations in subsequent iterations, even though examples of control actions leading to constraint violation are available to the controller.
This result further justifies the importance of learning-based controllers that take into account the confidence level of their predictions.

In Fig.~\ref{fig:improvement}, convergence is reached around $B=7$ for both GP-based controllers, achieving good economic performance when compared to the full-model benchmark.
The final converged operating trajectory is similar to that from the full-model solution, as observed in Fig.~\ref{fig:economic_2_cc} and Fig.~\ref{fig:solution}.
We observe that the shaded region also shrinks in the first few batches, in both final product mass and constraint violation, supporting the convergence of the system.
As discussed in Subsection~\ref{ssec: tracking results}, by obtaining more data around the optimal operating trajectory, the GP-MLMPC is able to take increasingly aggressive actions under certainty, leading to iterative improvements.
By mostly collecting samples around the optimal trajectory, the GP-MLMPC remains sample-efficient in converging towards the optimum.
Therefore, the results indicate that the proposed GP-MLMPC scheme is data efficient in learning an ideal controller for (semi-)batch processes.

It should also be noted that the initial PI trajectory is for the tracking task, and this is significantly different from the optimal trajectory for the economic task.
Therefore, this safe model learning approach enables a steady exploration of the feasible operating space, from a suboptimal trajectory that is far away from the optimal trajectory.
Results from the economic objective also indicate that the reaction rates are implicitly learned, in order to predict and optimize for product mass, while keeping the heat generated controlled.
These findings further support the GP-MLMPC as an effective learning scheme for optimal control.

Overall, the proposed GP-MLMPC scheme is able to achieve the same level of performance when compared to the full model benchmark by iterative improvement. 
Thus, this method offers the option of implementing eNMPC without an existing model of system dynamics to achieve satisfactory control in constrained batch processes.

\section{Conclusions and Outlook}\label{sec: conclusion}
The proposed GP-MLMPC scheme enables the adoption of NMPC without existing models and well-distributed data for the control of batch processes. 
The GP-MLMPC controller samples closely around the optimal trajectory while adhering to the tight safety thresholds.
Furthermore, we demonstrate empirically that GP-MLMPC is able to converge to optimal control within 8 iterations, highlighting the data efficiency of the approach.
With each batch iteration, the GP-MLMPC samples closer towards the optimal trajectory, thus reducing the need to sample across the whole operating space. 
This sample efficiency presents a critical advancement over established surrogate modeling approaches and points to a wide range of possible applications with low barriers. 
However, the convergence towards the optimal performance is not guaranteed by iterative sampling: a suboptimal trajectory may no longer provide new information to further iterations, resulting in the controller performing the same suboptimal trajectory, given little improvement in model accuracy.
Since the control objective is purely 'exploitative', there is no exploration mechanism by the controller to select actions that lead to probable improvement.
Thus, further work direction includes the investigation of active learning methods by further utilizing GP uncertainty quantification.

This work assumes that all system states are measurable with noise, while in real applications, variables such as species mass are usually not accessible in real time as measurements and require additional estimators.
Thus, the proposed approach should be extended to learn such estimators for more realistic systems.
Furthermore, this work is limited to the analysis of disturbance-free systems, where the only source of stochasticity is measurement noise.
Therefore, a further aim can also be to extend the approach to batch systems with disturbances.

\printcredits

\section*{Declaration of competing interest}
The authors declare that they have no known competing financial interests or personal relationships that could have appeared to influence the work reported in this paper.

\section*{Acknowledgments}
This project (GA number 101072732) has received funding from the HORIZON-MSCA-2021-DN-01 call of the research and innovation program of Horizon Europe 2021 under the Marie Skłodowska-Curie actions.

\section*{Appendix A. Log Marginal Likelihood for GP with SE Kernel}
For a GP with an SE kernel, the log marginal likelihood can be expressed in exact form in terms of $\Theta$~\cite{Rasmussen2006}:
\begin{equation*}\label{eq: log marginal likelihood}
\begin{aligned}
log~p(\bm{Y}|\bm{Z},\Theta) &= -\frac{1}{2}\bm{Y}^T(\bm{K}+\sigma_n^2I)^{-1}\bm{Y}\\
&- \frac{1}{2}log||\bm{K}+\sigma_n^2I||-\frac{n_D}{2}log\pi
\end{aligned}
\end{equation*}

\section*{Appendix B. Semi-batch Reactor Model}
We simulate the semi-batch polymerization model from~\cite{Lucia2014} as a case study.
We use the following ODEs for the simulation of the reactor:
\begin{subequations}
\begin{align*}
\left.\dfrac{dm_{\mathrm{W}}}{dt}\right|_t&= \dot{m}_{\mathrm{F}}(t) \omega_{\mathrm{W},\mathrm{~F}}\\
\left.\dfrac{dm_{\mathrm{A}}}{dt}\right|_t&= \dot{m}_{\mathrm{F}}(t) \omega_{\mathrm{A}, \mathrm{~F}}-k_{\mathrm{R} 1}(t) m_{\mathrm{A}, \mathrm{R}}(t)
\\&-k_{\mathrm{R} 2}(t) m_{\mathrm{AWT}} m_{\mathrm{A}}(t) / m_{\mathrm{ges}}(t)\\
\left.\dfrac{dm_{\mathrm{P}}}{dt}\right|_t&=  k_{\mathrm{R} 1}(t) m_{\mathrm{A}, \mathrm{R}}(t)+p_1 k_{\mathrm{R} 2}(t) m_{\mathrm{AWT}} m_{\mathrm{A}}(t) / m_{\mathrm{ges}}(t) \\\nonumber
\left.\dfrac{dT_{\mathrm{R}}}{dt}\right|_t&=  \frac{1}{c_{\mathrm{p}, \mathrm{R}} m_{\mathrm{ges}}(t)}\left[\dot{m}_{\mathrm{F}}(t) c_{\mathrm{p}, \mathrm{~F}}\left(T_{\mathrm{F}}-T_{\mathrm{R}}(t)\right)\right.\\
&+\Delta H_{\mathrm{R}} k_{\mathrm{R} 1}(t) m_{\mathrm{A}, \mathrm{R}}(t)-k_{\mathrm{K}}(t) A\left(T_{\mathrm{R}}(t)-T_{\mathrm{S}}(t)\right)\\
&\left.-\dot{m}_{\mathrm{AWT}} c_{\mathrm{p}, \mathrm{R}}\left(T_{\mathrm{R}}(t)-T_{\mathrm{EK}}(t)\right)\right] \\
\left.\dfrac{dT_{\mathrm{S}}}{dt}\right|_t&=  \frac{1}{c_{\mathrm{p}, \mathrm{~S}} m_{\mathrm{S}}}\left[k_{\mathrm{K}}(t) A\left(T_{\mathrm{R}}(t)-T_{\mathrm{S}}(t)\right)\right.\\
&\left.-k_{\mathrm{K}}(t) A\left(T_{\mathrm{S}}(t)-T_{\mathrm{M}}(t)\right)\right] \\
\left.\dfrac{dT_{\mathrm{M}}}{dt}\right|_t&=  \frac{1}{c_{\mathrm{p}, \mathrm{~W}} m_{\mathrm{M}, \mathrm{KW}}}\left[\dot{m}_{\mathrm{M}, \mathrm{KW}} c_{\mathrm{p}, \mathrm{~W}}\left(T_{\mathrm{M}}^{\mathrm{IN}}(t)-T_{\mathrm{M}}(t)\right)\right.\\
&\left.+k_{\mathrm{K}}(t) A\left(T_{\mathrm{S}}(t)-T_{\mathrm{M}}(t)\right)\right] \\\nonumber
\left.\dfrac{dT_{\mathrm{EK}}}{dt}\right|_t&=  \frac{1}{c_{\mathrm{p}, \mathrm{R}} m_{\mathrm{AWT}}}\left[\dot{m}_{\mathrm{AWT}} c_{\mathrm{p}, \mathrm{~W}}\left(T_{\mathrm{R}}(t)-T_{\mathrm{EK}}(t)\right)\right.\\
&-\alpha\left(T_{\mathrm{EK}}(t)-T_{\mathrm{AWT}}(t)\right) \\
&\qquad \left.+k_{\mathrm{R} 2}(t) m_{\mathrm{A}}(t) m_{\mathrm{AWT}} \Delta H_{\mathrm{R}} / m_{\mathrm{ges}}(t)\right] \\\nonumber
\left.\dfrac{dT_{\mathrm{AWT}}}{dt}\right|_t&=  {\left[\dot{m}_{\mathrm{AWT}, \mathrm{KW}} c_{\mathrm{p} \mathrm{~W}}\left(T_{\mathrm{AWT}}^{\mathrm{IN}}(t)-T_{\mathrm{AWT}}(t)\right)\right.} \\
&\qquad \left.-\alpha\left(T_{\mathrm{AWT}}(t)-T_{\mathrm{EK}}(t)\right)\right] /\left(c_{\mathrm{p}, \mathrm{~W}} m_{\mathrm{AWT}, \mathrm{KW}}\right)\\
\left.\dfrac{dT_{\mathrm{adiab}}}{dt}\right|_t&=\frac{\Delta H_{\mathrm{R}}}{m_{\mathrm{ges}}(t) c_{\mathrm{p}, \mathrm{R}}} \left.\dfrac{dm_{\mathrm{A}}}{dt}\right|_t\\
&-\left(\left.\dfrac{dm_{\mathrm{W}}}{dt}\right|_t +\left.\dfrac{dm_{\mathrm{A}}}{dt}\right|_t+\left.\dfrac{dm_{\mathrm{P}}}{dt}\right|_t\right)
\\&\cdot\left(\frac{m_{\mathrm{A}}(t) \Delta H_{\mathrm{R}}}{m_{\mathrm{ges}}^2(t) c_{\mathrm{p}, \mathrm{R}}}\right)+\left.\dfrac{dT_{\mathrm{R}}}{dt}\right|_t \\\nonumber
\text{where}\\
U(t)&=\frac{m_{\mathrm{P}}(t)}{m_{\mathrm{A}}(t)+m_{\mathrm{P}}(t)} \\
m_{\mathrm{ges}}(t)&=m_{\mathrm{W}}(t)+m_{\mathrm{A}}(t)+m_{\mathrm{P}}(t) \\
k_{\mathrm{R} 1}(t)&=k_0 e^{\frac{-E_a}{R\left(T_{\mathrm{R}}(t)+273.15\right)}}\left(k_{\mathrm{U} 1}(1-U(t))+k_{\mathrm{U} 2} U(t)\right) \\
k_{\mathrm{R} 2}(t)&=k_0 e^{\frac{-E_a}{R\left(T_{\mathrm{EK}}(t)+273.15\right)}}\left(k_{\mathrm{U} 1}(1-U(t))+k_{\mathrm{U} 2} U(t)\right) \\
k_{\mathrm{K}}(t)&=\left(m_{\mathrm{W}}(t) k_{\mathrm{WS}}+m_{\mathrm{A}}(t) k_{\mathrm{AS}}+m_{\mathrm{P}}(t) k_{\mathrm{PS}}\right) / m_{\mathrm{ges}}(t) \\
m_{\mathrm{A}, \mathrm{R}}(t)&=m_{\mathrm{A}}(t)-m_{\mathrm{A}}(t) m_{\mathrm{AWT}} / m_{\mathrm{ges}}(t) 
\end{align*}
\end{subequations}

\bibliographystyle{elsarticle-num}

\bibliography{references}

\begin{thebibliography}{10}
\expandafter\ifx\csname url\endcsname\relax
  \def\url#1{\texttt{#1}}\fi
\expandafter\ifx\csname urlprefix\endcsname\relax\def\urlprefix{URL }\fi
\expandafter\ifx\csname href\endcsname\relax
  \def\href#1#2{#2} \def\path#1{#1}\fi

\bibitem{Yoo2021}
H.~Yoo, H.~E. Byun, D.~Han, J.~H. Lee, {Reinforcement learning for batch process control: Review and perspectives}, Annual Reviews in Control 52 (2021) 108--119.
\newblock \href {https://doi.org/10.1016/J.ARCONTROL.2021.10.006} {\path{doi:10.1016/J.ARCONTROL.2021.10.006}}.

\bibitem{Lee2006}
J.~H. Lee, K.~S. Lee, {ITERATIVE LEARNING CONTROL APPLIED TO BATCH PROCESSES: AN OVERVIEW}, IFAC Proceedings Volumes 39~(2) (2006) 1037--1046.
\newblock \href {https://doi.org/10.3182/20060402-4-BR-2902.01037} {\path{doi:10.3182/20060402-4-BR-2902.01037}}.

\bibitem{Rawlings2017}
J.~B. Rawlings, D.~Q. Mayne, M.~Diehl, {Model predictive control : theory, computation, and design}, Nob Hill Publishing, 2017.

\bibitem{Lauri2014}
D.~Laur{\'{i}}, B.~Lennox, J.~Camacho, {Model predictive control for batch processes: Ensuring validity of predictions}, Journal of Process Control 24~(1) (2014) 239--249.
\newblock \href {https://doi.org/10.1016/J.JPROCONT.2013.11.005} {\path{doi:10.1016/J.JPROCONT.2013.11.005}}.

\bibitem{Lucia2014}
S.~Lucia, J.~A. Andersson, H.~Brandt, M.~Diehl, S.~Engell, {Handling uncertainty in economic nonlinear model predictive control: A comparative case study}, Journal of Process Control 24~(8) (2014) 1247--1259.
\newblock \href {https://doi.org/10.1016/j.jprocont.2014.05.008} {\path{doi:10.1016/j.jprocont.2014.05.008}}.

\bibitem{Fiedler2023}
F.~Fiedler, B.~Karg, L.~L{\"{u}}ken, D.~Brandner, M.~Heinlein, F.~Brabender, S.~Lucia, {do-mpc: Towards FAIR nonlinear and robust model predictive control}, Control Engineering Practice 140 (2023) 105676.
\newblock \href {https://doi.org/10.1016/J.CONENGPRAC.2023.105676} {\path{doi:10.1016/J.CONENGPRAC.2023.105676}}.

\bibitem{Marquez-Ruiz2019}
A.~Marquez-Ruiz, M.~Loonen, M.~B. Saltlk, L.~{\"{O}}zkan, {Model Learning Predictive Control for Batch Processes: A Reactive Batch Distillation Column Case Study}, Industrial and Engineering Chemistry Research 58~(30) (2019) 13737--13749.
\newblock \href {https://doi.org/10.1021/ACS.IECR.8B06474/ASSET/IMAGES/MEDIUM/IE-2018-06474P_0015.GIF} {\path{doi:10.1021/ACS.IECR.8B06474/ASSET/IMAGES/MEDIUM/IE-2018-06474P_0015.GIF}}.

\bibitem{Rasmussen2006}
C.~E. Rasmussen, C.~K.~I. Williams, Gaussian Processes for Machine Learning, The MIT Press, 2006.

\bibitem{Murray-Smith2003}
R.~Murray-Smith, D.~Sbarbaro, C.~E. Rasmussen, A.~Girard, {Adaptive, cautious, predictive control with gaussian process priors}, IFAC Proceedings Volumes 36~(16) (2003) 1155--1160.
\newblock \href {https://doi.org/10.1016/S1474-6670(17)34915-7} {\path{doi:10.1016/S1474-6670(17)34915-7}}.

\bibitem{Maciejowski2013}
J.~M. Maciejowski, X.~Yang, {Fault tolerant control using Gaussian processes and model predictive control}, Conference on Control and Fault-Tolerant Systems, SysTol (2013) 1--12\href {https://doi.org/10.1109/SYSTOL.2013.6693820} {\path{doi:10.1109/SYSTOL.2013.6693820}}.

\bibitem{Klenske2016}
E.~D. Klenske, M.~N. Zeilinger, B.~Sch{\"{o}}lkopf, P.~Hennig, {Gaussian Process-Based Predictive Control for Periodic Error Correction}, IEEE Transactions on Control Systems Technology 24~(1) (2016) 110--121.
\newblock \href {https://doi.org/10.1109/TCST.2015.2420629} {\path{doi:10.1109/TCST.2015.2420629}}.

\bibitem{Bradford2020}
E.~Bradford, L.~Imsland, D.~Zhang, E.~A. {del Rio Chanona}, {Stochastic data-driven model predictive control using gaussian processes}, Computers \& Chemical Engineering 139 (2020) 106844.
\newblock \href {http://arxiv.org/abs/1908.01786} {\path{arXiv:1908.01786}}, \href {https://doi.org/10.1016/J.COMPCHEMENG.2020.106844} {\path{doi:10.1016/J.COMPCHEMENG.2020.106844}}.

\bibitem{Hewing2020a}
L.~Hewing, J.~Kabzan, M.~N. Zeilinger, {Cautious Model Predictive Control Using Gaussian Process Regression}, IEEE Transactions on Control Systems Technology 28~(6) (2020) 2736--2743.
\newblock \href {http://arxiv.org/abs/1705.10702} {\path{arXiv:1705.10702}}, \href {https://doi.org/10.1109/TCST.2019.2949757} {\path{doi:10.1109/TCST.2019.2949757}}.

\bibitem{Maiworm2021}
M.~Maiworm, D.~Limon, R.~Findeisen, {Online learning-based model predictive control with Gaussian process models and stability guarantees}, International Journal of Robust and Nonlinear Control 31~(18) (2021) 8785--8812.
\newblock \href {http://arxiv.org/abs/1911.03315} {\path{arXiv:1911.03315}}, \href {https://doi.org/10.1002/RNC.5361} {\path{doi:10.1002/RNC.5361}}.

\bibitem{Micchelli2006}
C.~A. Micchelli, Y.~Xu, H.~Zhang, {Universal Kernels}, Journal of Machine Learning Research 7 (2006) 2651--2667.

\bibitem{Titsias2009}
M.~Titsias, Variational learning of inducing variables in sparse gaussian processes, in: D.~van Dyk, M.~Welling (Eds.), Proceedings of the Twelfth International Conference on Artificial Intelligence and Statistics, Vol.~5 of Proceedings of Machine Learning Research, PMLR, Hilton Clearwater Beach Resort, Clearwater Beach, Florida USA, 2009, pp. 567--574.

\bibitem{Mchutchon2011}
A.~Mchutchon, C.~E. Rasmussen, {Gaussian Process Training with Input Noise}, Advances in Neural Information Processing Systems 24 (2011).

\bibitem{Girard2002}
A.~Girard, C.~E. Rasmussen, J.~Q. Qui˜, Q.~Candela, R.~Murray-Smith, {Gaussian Process Priors with Uncertain Inputs Application to Multiple-Step Ahead Time Series Forecasting}, Advances in Neural Information Processing Systems 15 (2002).

\bibitem{Paulson2020}
J.~A. Paulson, E.~A. Buehler, R.~D. Braatz, A.~Mesbah, {Stochastic model predictive control with joint chance constraints}, International Journal of Control 93~(1) (2020) 126--139.
\newblock \href {https://doi.org/10.1080/00207179.2017.1323351} {\path{doi:10.1080/00207179.2017.1323351}}.

\bibitem{Hewing2020b}
L.~Hewing, K.~P. Wabersich, M.~Menner, M.~N. Zeilinger, {Learning-Based Model Predictive Control: Toward Safe Learning in Control}, Annual Review of Control, Robotics, and Autonomous Systems 3~(Volume 3, 2020) (2020) 269--296.
\newblock \href {https://doi.org/10.1146/ANNUREV-CONTROL-090419-075625/CITE/REFWORKS} {\path{doi:10.1146/ANNUREV-CONTROL-090419-075625/CITE/REFWORKS}}.

\bibitem{Silva2008}
B.~da~Silva, P.~Dufour, N.~Sheibat-Othman, S.~Othman, {Model Predictive Control of Free Surfactant Concentration in Emulsion Polymerization}, IFAC Proceedings Volumes 41~(2) (2008) 8375--8380.
\newblock \href {https://doi.org/10.3182/20080706-5-KR-1001.01416} {\path{doi:10.3182/20080706-5-KR-1001.01416}}.

\bibitem{Joy2019}
P.~Joy, K.~Rossow, F.~Jung, H.~U. Moritz, W.~Pauer, A.~Mitsos, A.~Mhamdi, {Model-based control of continuous emulsion co-polymerization in a lab-scale tubular reactor}, Journal of Process Control 75 (2019) 59--76.
\newblock \href {https://doi.org/10.1016/J.JPROCONT.2018.12.014} {\path{doi:10.1016/J.JPROCONT.2018.12.014}}.

\bibitem{Faust2021}
J.~M. Faust, S.~Hamzehlou, J.~R. Leiza, J.~M. Asua, A.~Mhamdi, A.~Mitsos, {Closed-loop in-silico control of a two-stage emulsion polymerization to obtain desired particle morphologies}, Chemical Engineering Journal 414~(October 2020) (2021) 128808.
\newblock \href {https://doi.org/10.1016/j.cej.2021.128808} {\path{doi:10.1016/j.cej.2021.128808}}.

\bibitem{Rostampour2015}
V.~Rostampour, P.~{Mohajerin Esfahani}, T.~Keviczky, {Stochastic Nonlinear Model Predictive Control of an Uncertain Batch Polymerization Reactor**This research was supported by the Netherlands Organization for Scientific Research (NWO) under the grant number 408-13-030.}, IFAC-PapersOnLine 48~(23) (2015) 540--545.
\newblock \href {https://doi.org/10.1016/j.ifacol.2015.11.334} {\path{doi:10.1016/j.ifacol.2015.11.334}}.

\bibitem{TempAccuracy}
{AN-021 - Determining Temperature Accuracy | Arroyo Instruments} (2026).

\bibitem{MassAccuracy}
{Types of Flow Sensors Used in Industrial Automation - Supmea Automation Co.,Ltd} (2026).

\bibitem{Nicholson2018}
B.~Nicholson, J.~D. Siirola, J.~P. Watson, V.~M. Zavala, L.~T. Biegler, pyomo.dae: a modeling and automatic discretization framework for optimization with differential and algebraic equations, Mathematical Programming Computation 10~(2) (2018) 187--223.
\newblock \href {https://doi.org/10.1007/S12532-017-0127-0/FIGURES/4} {\path{doi:10.1007/S12532-017-0127-0/FIGURES/4}}.

\bibitem{Wachter2006}
A.~W{\"{a}}chter, L.~T. Biegler, {On the implementation of an interior-point filter line-search algorithm for large-scale nonlinear programming}, Mathematical Programming 106~(1) (2006) 25--57.
\newblock \href {https://doi.org/10.1007/S10107-004-0559-Y/METRICS} {\path{doi:10.1007/S10107-004-0559-Y/METRICS}}.

\bibitem{gpy2014}
{GPy}, {GPy}: A gaussian process framework in python, \url{http://github.com/SheffieldML/GPy} (since 2012).

\bibitem{Andersson2019}
J.~A.~E. Andersson, J.~Gillis, G.~Horn, J.~B. Rawlings, M.~Diehl, {CasADi} -- {A} software framework for nonlinear optimization and optimal control, Mathematical Programming Computation 11~(1) (2019) 1--36.
\newblock \href {https://doi.org/10.1007/s12532-018-0139-4} {\path{doi:10.1007/s12532-018-0139-4}}.

\bibitem{Ziegler1942}
J.~G. Ziegler, N.~B. Nichols, {Optimum Settings for Automatic Controllers}, Journal of Fluids Engineering 64~(8) (1942) 759--765.
\newblock \href {https://doi.org/10.1115/1.4019264} {\path{doi:10.1115/1.4019264}}.

\bibitem{Myren2021}
S.~Myren, E.~Lawrence, {A comparison of Gaussian processes and neural networks for computer model emulation and calibration}, Statistical Analysis and Data Mining 14~(6) (2021) 606--623.
\newblock \href {https://doi.org/10.1002/SAM.11507;WGROUP:STRING:PUBLICATION} {\path{doi:10.1002/SAM.11507;WGROUP:STRING:PUBLICATION}}.

\bibitem{Goldin2023}
M.~A. Goldin, S.~Virgili, M.~Chalk, {Scalable Gaussian process inference of neural responses to natural images}, Proceedings of the National Academy of Sciences of the United States of America 120~(34) (2023) e2301150120.
\newblock \href {https://doi.org/10.1073/PNAS.2301150120;WEBSITE:WEBSITE:PNAS-SITE;WGROUP:STRING:PUBLICATION} {\path{doi:10.1073/PNAS.2301150120;WEBSITE:WEBSITE:PNAS-SITE;WGROUP:STRING:PUBLICATION}}.

\end{thebibliography}

\end{document}